\documentclass[runningheads]{llncs}
\usepackage{graphicx}

\usepackage{tikz}
\usepackage{comment}
\usepackage{amsmath,amssymb}
\usepackage{booktabs}
\usepackage{cite}
\usepackage{mathtools}
\usepackage{microtype}
\usepackage{multicol}
\usepackage{caption}
\usepackage{orcidlink}
\usepackage{subcaption}
\usepackage{xspace}
\usepackage{xcolor} 
\definecolor{lblue}{RGB}{0, 112, 192}

\usepackage[accsupp]{axessibility}  

\newcommand\msemin{$\text{MSE}_{\min}$\xspace}
\newcommand\fmnist{Fusion\-MNIST\xspace}
\newcommand\fceleba{Fusion\-CelebA\xspace}
\newcommand\ftless{Fusion\-T-LESS\xspace}
\newcommand\fvae{Fusion\-VAE\xspace}
\newcommand{\bbs}[1]{\boldsymbol #1}
\newcommand{\bbx}{\boldsymbol x}
\newcommand{\bby}{\boldsymbol y}
\newcommand{\bbz}{\boldsymbol z}
\newcommand{\tb}{\textbf}

\newif\ifsupplementaryheader
\supplementaryheadertrue

\usepackage[capitalize]{cleveref}
\crefname{section}{Sec.}{Secs.}
\Crefname{section}{Section}{Sections}
\Crefname{table}{Table}{Tables}
\crefname{table}{Tab.}{Tabs.}

\begin{document}

\pagestyle{headings}
\mainmatter
\def\ECCVSubNumber{5001}

\title{FusionVAE: A Deep Hierarchical Variational Autoencoder for RGB Image Fusion}

\titlerunning{FusionVAE: A Deep Hierarchical Variational Autoencoder}

\author{Fabian Duffhauss\inst{1,2}\orcidlink{0000-0002-8910-3852} 
\and
Ngo Anh Vien\inst{1}\orcidlink{0000-0001-9646-267X} 
\and
Hanna Ziesche\inst{1}\orcidlink{0000-0003-2042-3660} 
\and
Gerhard~Neumann\inst{3}\orcidlink{0000-0002-5483-4225}
}
\authorrunning{F. Duffhauss et al.}
%
\institute{Bosch Center for Artificial Intelligence\\
\email{\{Fabian.Duffhauss,AnhVien.Ngo,Hanna.Ziesche\}@bosch.com} \and
University of Tübingen \and
Karlsruhe Institute of Technology\\
\email{gerhard.neumann@kit.edu}}

\maketitle


\begin{abstract}

Sensor fusion can significantly improve the performance of many computer vision tasks. 
However, traditional fusion approaches are either not data-driven and cannot exploit prior knowledge nor find regularities in a given dataset or they are restricted to a single application.
We overcome this shortcoming by presenting a novel deep hierarchical variational autoencoder called \fvae that can serve as a basis for many fusion tasks. 
Our approach is able to generate diverse image samples that are conditioned on multiple noisy, occluded, or only partially visible input images. We derive and optimize a variational lower bound for the conditional log-likelihood of \fvae. In order to assess the fusion capabilities of our model thoroughly, we created three novel datasets for image fusion based on popular computer vision datasets. In our experiments, we show that \fvae learns a representation of aggregated information  that is relevant to fusion tasks. The results demonstrate that our approach outperforms traditional methods significantly. Furthermore, we present the advantages and disadvantages of different design choices.

\end{abstract}

\section{Introduction}

Sensor fusion is a popular technique in computer vision as it allows to combine the advantages from multiple information sources. It is especially gainful in scenarios where a single sensor is not able to capture all necessary data to perform a task satisfactorily. Over the last years, we have seen many examples, where the accuracy of computer vision tasks was significantly improved by sensor fusion, e.g.\ in environmental perception for autonomous driving \cite{mv3d, avod, 3d_cvf}, for 6D pose estimation \cite{densefusion, pvn3d, ffb6d}, and for robotic grasping \cite{zhang2017robust, graspfusionnet}. However, traditional fusion methods usually focus more on the beneficial merging of multiple modalities and less on teaching the model to obtain profound prior knowledge about the used dataset. 

\begin{figure}[t]
  \centering
  \includegraphics[page=1, trim = 48mm 43mm 75.5mm 48.5mm, clip,  width=0.6\linewidth]{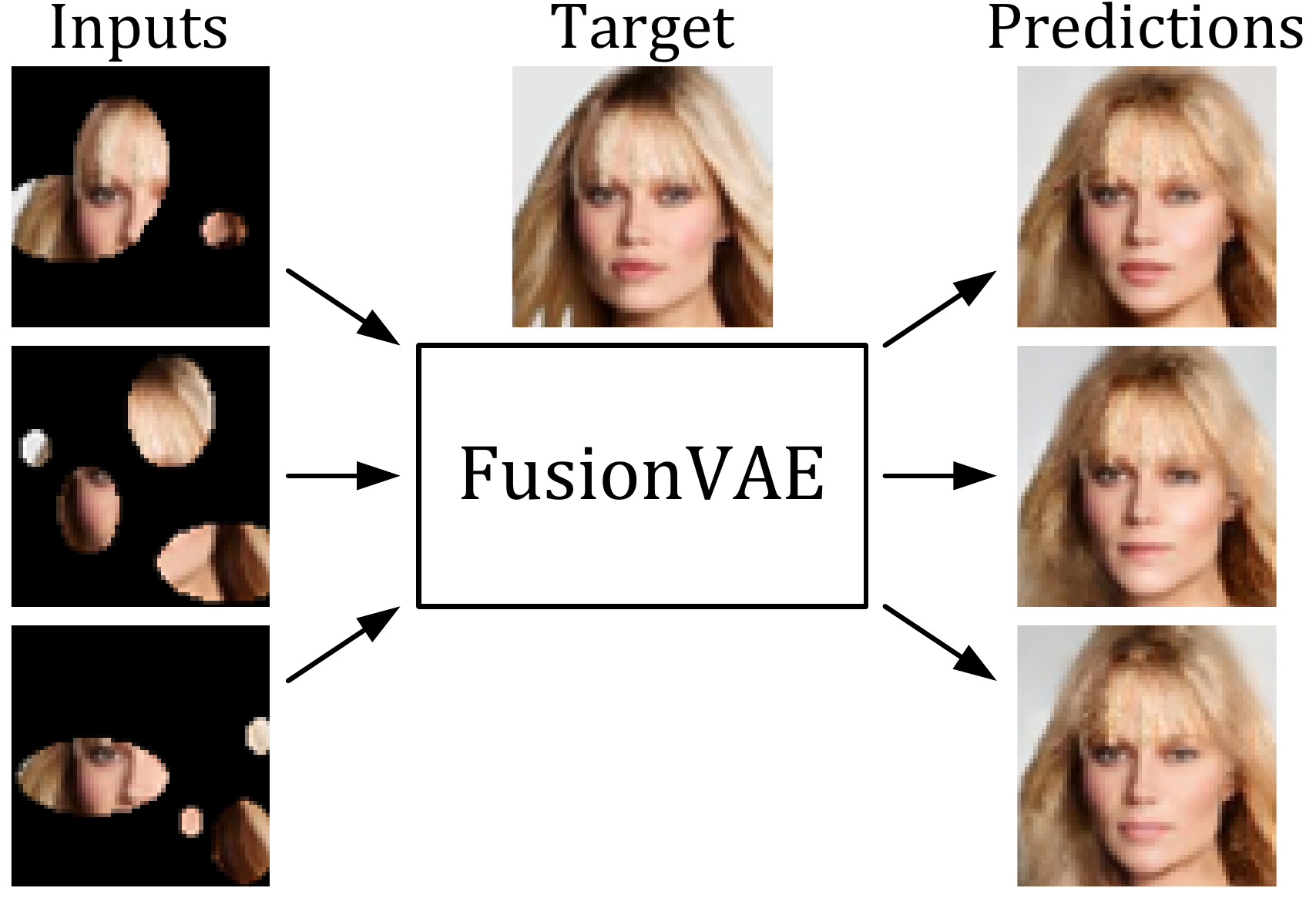}
   \caption{Overview of our \fvae approach. The network receives up to three partly occluded input images, fuses them together with prior knowledge, and predicts different hypothesis of how the target images could look like.}
   \label{fig_fvae_overview}
\end{figure}

Our work tries to fill in this research gap by proposing a deep hierarchical variational autoencoder called \fvae that is able to perform both tasks: fusing information from multiple sources and supplementing it with prior knowledge about the data gained while training. As shown in \cref{fig_fvae_overview}, \fvae merges a varying number of input images for reconstructing the original target image using prior knowledge about the dataset. To the best of our knowledge, \fvae is the first approach that combines these two tasks. Therefore, we developed three challenging benchmarks based on well-known computer vision datasets to evaluate the performance of our approach. In addition, we perform comparisons to baselines by extending traditional approaches to perform the same tasks. \fvae outperforms all these traditional methods on all proposed benchmark tasks significantly. We show that \fvae can generate high-quality images given few input images with partial observability. We provide ablation studies to illustrate the impact of commonly used information aggregation operations and to prove the benefits of the employed posterior distribution.

We can summarize the four main contributions of our paper as follows: i)~We create three challenging image fusion tasks for generative models. ii)~We develop a deep hierarchical VAE called \fvae that is able to perform image-based data fusion while employing prior knowledge of the used dataset. iii)~We show that \fvae produces high-quality fused output images and outperforms traditional methods by a large margin. iv) We perform ablation studies showing the benefits of our design choices regarding both the posterior distribution and commonly used aggregation methods.

\section{Related Work}

In this section, we present related work about image generation, image fusion, and image completion.

\subsection{VAE-based Image Generation}

Variational autoencoders (VAE) are powerful networks that are able to compress the essence of datasets in a small latent space while being able to exploit it for generative tasks \cite{vae}. However, the standard VAE is limited in capacity and expressiveness and thus, when applied to image generation leads to over-smoothed results lacking fine-grained details. Over the last years much work has been invested into the effort of improving the generative performance of VAEs.
One stream of work is based on introducing a hierarchy into the latent space of the VAE and scaling this hierarchy to greater and greater depth. First introduced in \cite{lvae} many hierarchical VAEs are based on coupling the inference and generative processes by introducing a deterministic bottom-up path combined with a stochastic top-down process in the inference network and sharing the latter with the generative model. This setting has been extended by an additional deterministic top-down path and bidirectional inference in \cite{biva}. Recently, very deep hierarchical VAEs were realized in \cite{veryDeepVAE} by introducing  residual bottlenecks with dedicated scaling, update skipping, and nearest neighbour up-sampling. Closest to our work is the recently proposed NVAE architecture \cite{nvae}, which relies on depth-wise convolution, residual posterior parametrization, and spectral regularization to enhance stability. 

Other approaches propose increasing the expressiveness of VAEs by combining them with auto-regressive models like RNNs or PixelCNNs \cite{ChenKSDDSSA16, pixelVAE, pixelVAE2, draw}, conditioning contexts (CVAE) \cite{cvae,walker2016uncertain}, normalizing flows \cite{KingmaSW16}, generative adversarial networks (GANs) \cite{LarsenSW16, parmar2021dual}, or variational generative adversarial networks (CVAE-GAN) \cite{bao2017cvae}.

\subsection{Fusion of Multiple Images}

Image fusion has long been dominated by classical computer vision. Only lately deep learning methods entered the domain with the CNN-based approach proposed by Liu et al. \cite{liu2017multi}. In a subsequent publication the authors extended their work to a multi-scale setting \cite{liu2017medical}. Shortly afterwards, Prabhakar et al. developed a fusion method based on a siamese network architecture, called DeepFuse \cite{DeepFuse} which was improved in subsequent work \cite{DenseFuse} by employing the DenseNet architecture \cite{DenseNet}.  
Concurrently, Li et al. \cite{liVGG} proposed a fusion architecture based on VGG \cite{vgg} and in order to scale to even greater depth another one \cite{li2019infrared} based on ResNet-50 \cite{resnet}. 
The aforementioned methods use CNNs as feature extractors and as decoders, while the fusion operations themselves are restricted to classical methods like averaging or addition of feature maps or weighted source images. 
A fully CNN-based feature-map fusion mechanism was proposed in \cite{unsupervisedDIF}. 

While all previous publications target only a specific fusion task (e.g. multi-focus fusion, multi-resolution fusion, etc.) or were limited to specific domains (e.g. medical images), two very recent works propose novel multi-purpose fusion networks, which are applicable to many fusion tasks and image types \cite{fusionDN, rethinking}. Very recently also GAN-based methods entered the domain of image fusion, starting with the work by Ma et al. on infrared-visible fusion \cite{ma2020infrared, FusionGAN} and with \cite{DDcGAN, Xu2019LearningAG} on multi-resolution image fusion. Most recent are two publications on GAN-based multi-focus image fusion \cite{FuseGAN, huang2020gan}.
While GAN-based approaches can generate high-fidelity images, it is known that they suffer from the mode collapse problem. VAE-based methods in contrast are known to generate more faithful data distribution \cite{nvae}. Different from previous work, this paper proposes a VAE-based multi-purpose fusion framework.

\subsection{Image Completion}

Similar to image fusion, also image completion has only recently become a playing field for deep learning methods. First approaches based on simple multilayer perceptrons (MLPs) \cite{kohler2014mask} or CNNs \cite{gu2018cnn} were targeted only to filling small holes in an image. However, with the introduction of GANs \cite{gan}, the area quickly became dominated by GAN-based approaches, starting with the context encoders presented by Pathak et al. \cite{2016contextEncoders}. Many subsequent papers proposed extensions to this model in order to obtain fine-grained completions while preserving global coherence by introducing additional discriminators \cite{iizuka2017globally}, searching for closest samples to the corrupted image in a latent embedding space conditioning on semantic labels \cite{2018spgNet}, or designing additional specialized loss functions \cite{2017generativeFace}. High resolution results were obtained recently by multi-scale approaches \cite{yang2017high}, iterative upsampling \cite{zeng2020high}, and the application of contextual attention \cite{yu2018generative, song2018contextual, yan2018shiftNet}. 
Another stream of current work focuses on multi-hypothesis image completion, leveraging probabilistic problem formulations \cite{zheng2019pluralistic, marinescu2020bayesian}.

\section{Background}

In this section, we outline the fundamentals of standard VAEs, conditional VAEs, and hierarchical VAEs upon which we build our approach. 
Another section is dedicated to aggregation methods for data fusion.

\subsection{Standard VAE}

A variational autoencoder (VAE) \cite{vae} is a neural network consisting of a probabilistic encoder $q(\bbz|\bby)$ and a generative model $p(\bby|\bbz)$. The generator models a distribution over the input data $\bby$, conditioned on a latent variable $\bbz$  with  prior distribution $p_{\theta}(\bbz)$.
The encoder approximates the posterior distribution $p(\bbz|\bby)$ of the latent variables $\bbz$ given input data $\bby$ and is trained along with the generative model by maximizing the evidence lower bound (ELBO)
\begin{equation}
\label{eq_VLB}
    {\rm ELBO}(\bby) = \mathbb{E}_{q(\boldsymbol{z} | \bby)} 
    [\log p(\bby | \boldsymbol{z})] 
    - \text{KL}( q(\bbz | \bby) || p(\bbz) ),
\end{equation}
where KL is the Kullback–Leibler divergence and $\log p(\bby) \ge {\rm ELBO}(\bby)$.

\subsection{Conditional VAE}

In VAEs, the generative model $p(\bby|\bbz)$ is unconditional. In contrast, conditional VAEs (CVAE)  \cite{cvae} consider a generative model for a conditional distribution $p(\bby|\bbx,\bbz)$ where $\bby$ is the target data, $\bbx$ is the conditional input variable, and $\bbz$ is a latent variable. The prior of the latent variable is $p(\bbz|\bbx)$, while its approximate posterior distribution is given by $q(\bbz|\bbx,\bby)$. The variational lower bound of the conditional log-likelihood can be written as follows
\begin{equation}
    \log p(\bby|\bbx) \ge \mathbb{E}_{q(\bbz | \bbx,\bby)} 
    [\log p(\bby |\bbx, \bbz)] 
    - \text{KL}( q(\bbz | \bbx,\bby) || p(\bbz|\bbx) ).
\end{equation}

\subsection{Hierarchical VAE}

In hierarchical VAEs \cite{lvae, nvae, SetVAE}, the latent variables $\bbz$ are divided into $L$ disjoint groups $\bbz_1, ..., \bbz_L$ in order to increase the expressiveness of both prior and approximate posterior which become
\begin{equation}
    p(\bbz) = \prod_{l=1}^L p(\bbz_l | \bbz_{<l})
\quad \text{and} \quad
    q(\bbz | \bby) = \prod_{l=1}^L q(\bbz_l | \bbz_{<l}, \bby),
\end{equation}
where $\bbz_{<l}$ denotes the latent variables in all previous hierarchies.
All the conditionals in the prior $p(\bbz_l | \bbz_{<l})$ and in the approximate posterior  $q(\bbz_l | \bbz_{<l}, \bby)$ are modeled by factorial Gaussian distributions.
Under this modelling choice, the ELBO from \cref{eq_VLB} turns into
\begin{equation}
\label{eq_hierVLB}
    {\rm ELBO}(\bby) = \mathbb{E}_{q(\bbz | \bby)} 
    [\log p(\bby | \bbz)]
    - \sum_{l=1}^L \mathbb{E}_{q(\bbz_{<l} | \bby)}
    [\text{KL}( q(\bbz_l | \bbz_{<l}, \bby) 
    || p(\bbz_l|\bbz_{<l}) )].
\end{equation}

\subsection{Aggregation Methods}
\label{sec_aggregation_methods}

For fusing data within our approach, we consider different aggregation methods, such as mean aggregation, max aggregation, Bayesian aggregation \cite{baco} and pixel-wise addition. All described aggregation methods fuse a set of feature tensors $\boldsymbol f_1, ...,\boldsymbol f_K$, obtained by encoding $K$ input images $\{\bbx_i\}_{i=1}^K$ in a permutation invariant way \cite{ZaheerKRPSS17}.
In mean aggregation, multiple feature vectors are fused by taking the pixel-wise average $\boldsymbol f = \frac{1}{K} \sum^K_{i=1} \boldsymbol f_i$.
For max aggregation, we take the pixel-wise maximum instead $\boldsymbol f = \max_i(\boldsymbol f_i)$.
Bayesian aggregation (BA) \cite{baco} considers an uncertainty estimate for the fused feature vectors. In order to obtain such an uncertainty estimate, the encoder has to predict means $\boldsymbol \mu_i$ and variances $\boldsymbol \sigma_i$ of a factorized Gaussian distribution over the latent feature vectors
\begin{equation}
    \boldsymbol f_i = \mathcal{N}(\boldsymbol \mu_i, \text{diag}(\boldsymbol \sigma_i)), 
    \text{with~} \boldsymbol \mu_i=\text{enc}_\mu(\bbx_i, \bby) \text{~and~} \boldsymbol \sigma_i=\text{enc}_\sigma(\bbx_i, \bby),
\end{equation}
instead of $\boldsymbol f_i$ directly.
Here $\text{enc}_\mu$ and $\text{enc}_\sigma$ represent the encoding process which generates means and variances respectively. The predicted distributions over latent feature vectors for multiple input images can be fused iteratively using the Bayes rule \cite{rkn} (detailed derivation is given in \cref{sec_derivation_of_bayesian_aggregation})
\begin{equation}
    \boldsymbol q_i = \boldsymbol \sigma^2_{i-1} \oslash (\boldsymbol \sigma^2_{i-1} + \boldsymbol \sigma^2_i)
\end{equation}
\begin{equation}
    \text{where~} 
    \boldsymbol \mu_i = \boldsymbol \mu_{i-1} + \boldsymbol q_i \odot (\boldsymbol \mu_i - \boldsymbol \mu_{i-1})
\quad     \text{and~} \quad
    \boldsymbol \sigma^2_i = \boldsymbol \sigma^2_{i-1} \odot (1 - \boldsymbol q_i).
\end{equation}
$\oslash$ and $\odot$ denote element-wise division and multiplication respectively.

\section{Conditional Generative Models for Image Fusion}

We propose a deep hierarchical conditional variational autoencoder, called \fvae \emph{(Fusion Variational Auto-Encoder)}, that is able to fuse information from multiple sources and to infer the missing information in the images from a prior learned from the dataset. To the best of our knowledge, it is the first model that combines the generative ability of a hierarchical VAE to learn the underlying distribution of complex datasets with the ability to fuse multiple input images.

\subsection{Problem Formulation}

We consider image fusion problems that are concerned with generating the fused target image from multiple source images. Each source image contains partial information of the target image and the goal of the task is to recover the original target image given a finite set of source images. In particular, we denote the target image as $\bby$ and the set of $K$ source images as context $\bbx=\{\bbx_1, \,\bbx_2, \,\ldots, \,\bbx_K\}$, where each $\bbx_i$ is one source image. Given training sample $(\bbx, \bby)$, we aim to maximize the conditional likelihood $p(\bby|\bbx)$.

\subsection{\fvae}

Our approach is designed to maximize the conditional likelihood $p(\bby|\bbx)$. However, optimizing this objective directly is intractable. Therefore, we derive a variational lower bound as follows (detailed derivation can be found in Appendix \ref{elbo_derivation})
\begin{align}
\begin{split}
    \log p(\bby|\bbx) 
    &\ge \mathbb{E}_{q_{\phi}(\bbz |\bby)} 
    [\log p_{\theta}(\bby |\bbx, \bbz)] \\
    &- \beta \sum_{l=1}^L \alpha_l \mathbb{E}_{q(\bbz_{<l} |\bby)}
    [\text{KL}( q_{\phi}(\bbz_l |\bby, \bbz_{<l}) 
    || p_{\theta}(\bbz_l| \bbx, \bbz_{<l}) )],
\end{split}
\end{align}
where we split the latent variables $\bbz$ into $L$ disjoint groups $\{\bbz_1, \,\bbz_2, \,\ldots, \,\bbz_L\}$. 
$\beta$ and $\alpha_l$ are annealing parameters that control the warming-up of the KL terms as in \cite{nvae}. 
Inspired by \cite{lvae}, $\beta$ is increased linearly from 0 to 1 during the first few training epochs to start training the reconstruction before introducing the KL term, which is increased gradually.
$\alpha_l$ is a KL balancing coefficient \cite{dvaepp} that is used during the warm-up period to encourage the equal use of all latent groups and to avoid posterior collapse. 

\begin{figure}[t]
  \centering 
  \includegraphics[page=1, trim = 11mm 47mm 8mm 43mm, clip, width=1.0\linewidth]{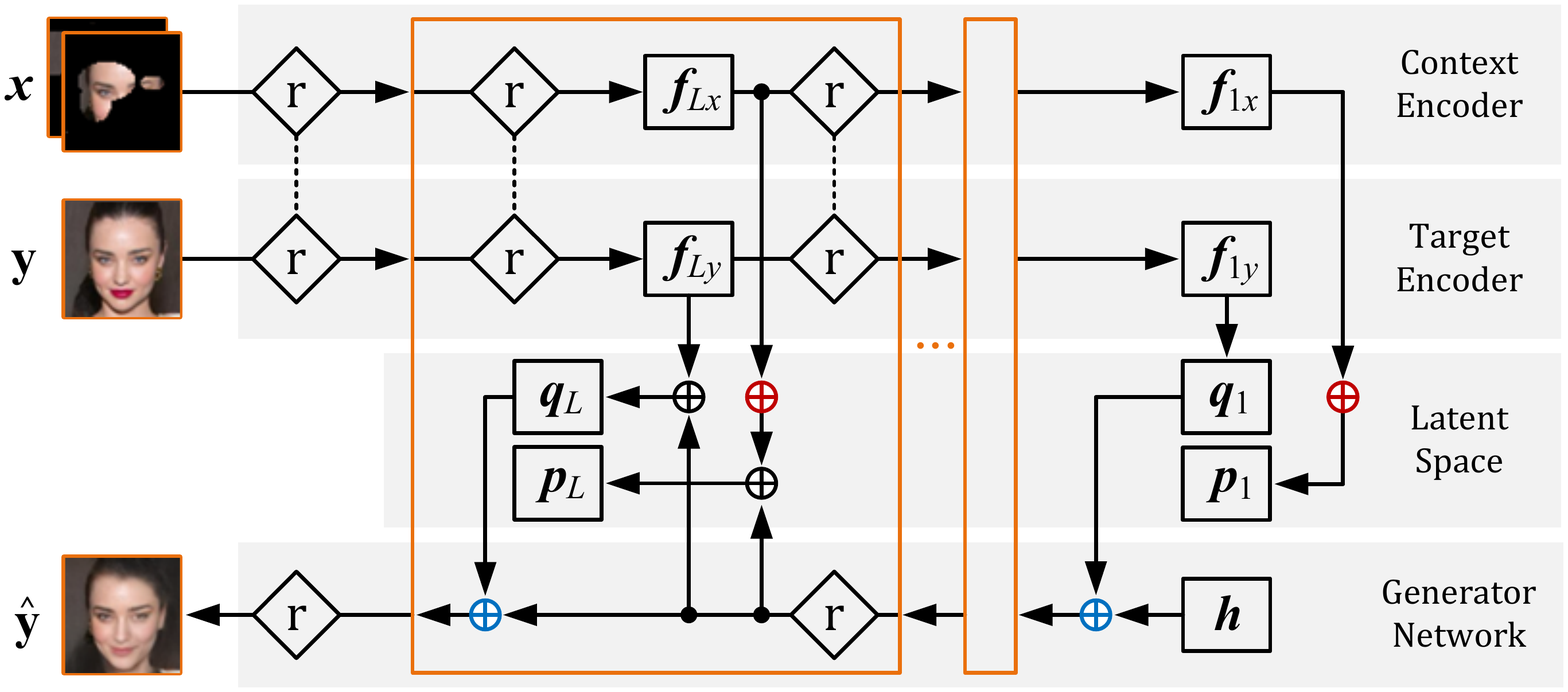}
   \caption{Overview of the proposed network architecture.
   $\boldsymbol h$ is a trainable parameter vector, \textcolor{lblue}{$\boldsymbol{\oplus}$} denotes concatenation, \textcolor{red}{$\boldsymbol{\oplus}$} max aggregation, and $\boldsymbol{\oplus}$ pixel-wise addition. \includegraphics[page=1, trim = 6.5mm 6.4mm 6.5mm 6.5mm, clip,  scale=0.5]{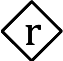} is a residual network like in \cite{nvae}. The dotted lines between the residual networks indicate shared parameters.}
   \label{fig_our_architecture}
\end{figure}

\fvae consists of three main networks and a latent space as illustrated in \cref{fig_our_architecture}: 1) a context encoder network which models the conditional prior $p_{\phi}(\bbz_l| \bbx, \bbz_{<l})$, 2) a target encoder which models the approximate posterior $q_{\phi}(\bbz | \bby)$, 3) a latent space comprising the $L$ latent groups, and 4) a generator network $p_{\theta}(\bby |\bbx, \bbz)$ that aims to reconstruct the target image.

\subsection{Network Architecture}

\cref{fig_our_architecture} illustrates the network architecture of our \fvae for training. It is built in a hierarchical way inspired by \cite{nvae}.
In each latent hierarchy $l \in 1, \,\ldots, \,L$
we have a set of feature maps $\boldsymbol f_{lx}$, $\boldsymbol f_{ly}$ and latent distributions $p_l$, $q_l$.

The first gray box contains the context encoder network that obtains a stack of source images $\bbx$ and employs residual cells \cite{resnet} as in \cite{nvae} to extract features $\boldsymbol f_{lx}$.
The second gray box shows the target encoder network that encodes the target image $\bby$ into the feature map $\boldsymbol f_{ly}$ using the same residual cells as the context encoder. 
The third gray box illustrates the latent space which contains the prior distributions $p_{\phi}(\bbz_l| \bbx, \bbz_{<l})$ (denoted $p_1, \,..., \,p_L$ in \cref{fig_our_architecture}) and the approximate posterior distributions $q_{\phi}(\bbz_l |\bby, \bbz_{<l})$ (denoted $q_1, \,..., \,q_L$ in \cref{fig_our_architecture}).
The fourth gray box contains the generator network which aims to create different output samples $\boldsymbol{\hat{y}}$. It employs a trainable parameter vector $\boldsymbol h$, concatenates the information from all hierarchies, and decodes them using residual cells.

In each latent hierarchy, we aggregate the context features $\boldsymbol f_{lx}$ using pixel-wise max aggregation.
In all but the first hierarchy, we pixel-wisely add the corresponding feature map from the generator network to the aggregated context features and to the target image features $f_{ly}$. Using 2D convolutional layers, we learn the prior distributions $p_l$ and the approximate posterior distributions $q_l$.
We propose to use the approximate posterior distributions $q_l$ as target distributions in order to learn good prior distributions $p_l$. Therefore, $q_{\phi}(\bbz_l |\bby, \bbz_{<l}) $ is created from the target image features $f_{ly}$ as well as information from the generator network. 

During training the generator network aims to create a prediction $\boldsymbol{\hat{y}}$ based on samples of the posterior distributions $q_l$ and a trainable parameter vector $\boldsymbol h$.

For evaluation, we can omit the target image input $\bby$ and sample from the prior distributions $p_l$.
In case no input image is given, we set $p_1$ to a standard normal distribution. Based on the samples and the trainable parameter vector $\boldsymbol h$, our \fvae can generate new output images.

\section{Experimental Setup}
\label{sec:experimentalSetup}

To evaluate our approach, we conduct a series of experiments on three different datasets using data augmentation. Furthermore, we adapt traditional architectures for solving the same tasks in order to compare our results. Finally, we perform an ablation study to show the effects of specific design choices.

\subsection{Datasets}

For training and evaluating our approach, we create three novel fusion datasets based on MNIST \cite{mnist}, CelebA \cite{celeba}, and T-LESS \cite{tless} as described in the following.

\paragraph{FusionMNIST.}

Based on the dataset MNIST \cite{mnist}, we create an image fusion dataset called \fmnist. For each target image, it contains different noisy representations where only random parts of the target image are visible. 
The first three columns of \cref{fig_mnist_results} show different examples of \fmnist corresponding to the target images in the fourth column. To generate \fmnist, we applied zero padding to all MNIST images to obtain a resolution of $32 \times 32$. For creating a noisy representation, we generate a mask composed of the union of a varying number of ellipses with random size, shape and position. All parts of the given images outside the mask are blackened. Finally, we add Gaussian noise with a fixed variance and clip the pixel values afterwards to stay within $[0, 1]$.

\paragraph{FusionCelebA.}

We generate a similar fusion dataset based on the aligned and cropped version of CelebA \cite{celeba} which we call \fceleba. \cref{fig_celeba_results} depicts different example images in the first three columns which belong to the target image in the fourth column. 
To generate \fceleba, we center-crop the CelebA images to $148 \times 148$ before scaling them down to $64 \times 64$ as proposed by \cite{LarsenSW16}.
As in \fmnist, we create different representations by using masks based on random ellipses.

\paragraph{FusionT-LESS.}

A promising area of application for our \fvae is robot vision. Scenes in robotics settings can be very difficult to understand due to texture-less or reflective objects and occlusions. To examine the performance of our \fvae in this area, we create an object dataset with challenging occlusions based on T-LESS \cite{tless} which we call \ftless. To generate \ftless, we use the real training images of T-LESS and take all images of classes 19 -- 24 as basis for the target images. This selection contains all objects with power sockets and therefore images with many similarities.
Every tenth image is removed from the training set and used for evaluation. 
In order to create challenging occlusions, we cut all objects from images of other classes using a Canny edge detector \cite{canny} and overlay each target image with a random number between five and eight cropped objects. We select all images from classes 1, 2, 5 -- 7, 11 -- 14, and 25 -- 27 as occluding objects for training and classes 3, 4, 8 -- 10, 15 -- 18, and 28 -- 30 for evaluation.

\subsection{Data Augmentation}
\label{sec_augmentations}

During training we apply different augmentation methods on the datasets to avoid overfitting. For \fmnist, we apply the elliptical mask generation and the addition of Gaussian noise live during training so that we obtain an infinite number of different fusion tasks. For \ftless, almost the entire creation of occluded images is performed during training. We apply horizontal flips, rotations, scaling and movement of target and occluding images with random parameters before composing the different occluded representations. Solely the object cutting with the Canny edge detector is performed offline as a pre-processing step to keep the training time low. For \fceleba, we apply a horizontal flip of all images randomly in 50\% of all occasions and also the elliptical mask generation is done live during training.

\subsection{Architectures for Comparison}
\label{sec_architectures_for_comparison}

To the best of our knowledge, \fvae is the first fusion network for multiple images with a generative ability to fill areas without input information based on prior knowledge about the dataset under consideration. 
For lack of a suitable other model from the literature which would allow a fair comparison on our multi-image fusion tasks, we compare our approach with standard architectures that we adapted to support our tasks.

The first architecture for comparison is a CVAE with residual layers as employed in \cite{nvae}. We use a shared encoder for processing the input images and applied max aggregation before the latent space as we did in our \fvae.
The second architecture for comparison is a fully convolutional network (FCN) with shared encoder and max aggregation before the decoder.

For both baseline architectures, we created a version with skip connections (+S) and a version without. When using skip connections, we applied max aggregation at each shortcut for merging the features from the encoder with the decoder's features. To allow for a fair comparison, we designed all architectures so that they have a similar number of trainable parameters.

\subsection{Training Procedure}

We trained all networks in a supervised manner using the augmented target images $\bby$ as described in \cref{sec_augmentations}. In order to teach the networks both to fuse information from a different number of input images and to learn prior knowledge about the dataset, we vary the number of input images $\bbx$ during the entire training. Specifically, we select a uniformly distributed random number between zero and three for each batch.

\subsection{Evaluation Metrics}

For evaluation, we estimate the negative log-likelihood (NLL) in bits per dimension (BPD) using weighted importance sampling \cite{importanceWeight}. We use 100 samples for all experiments with \fceleba as well as \ftless and 1000 samples for \fmnist. Since we cannot estimate the NLL of the FCN, we used the minimum over all samples of the mean squared error (\msemin) as second metric. 

\section{Results}
\label{sec:results}

This section presents and discusses the quantitative and qualitative results of our research in comparison to the baseline methods mentioned in \cref{sec_architectures_for_comparison}.

\subsection{Quantitative Results}

\cref{tab_fmnist,tab_fceleba,tab_ftless} show the NLL and the \msemin of all architectures on \fmnist, \fceleba, and \ftless respectively. The results are divided into the results based on zero to three input images and the average (avg) of it. We see that our \fvae outperforms all baseline methods on average.
Regarding the NLL, our model surpasses the others additionally for 0 and 1 input images. For 2 and 3 images, 
CVAE+S reaches sometimes slightly better NLL values. However, our approach reaches the best \msemin values for each number of input images.

\begin{table}[h]
	\centering
	\tabcolsep=0.1cm
\begin{tabular}{l|ccccc|ccccc}
	\toprule
      &        \multicolumn{5}{c}{NLL in $10^{-2}$ BPD}       &     \multicolumn{5}{|c}{\msemin in $10^{-2}$}   \\ 
      &    0      &    1     &    2     &    3     &    avg   &    0    &    1    &    2    &    3    &    avg  \\
     \midrule
FCN   &           &          &          &          &          &   10.99 &    5.81 &    5.78 &    5.79 &    7.25 \\
FCN+S &           &          &          &          &          &    5.80 &    3.74 &    2.54 &    1.78 &    3.56 \\
CVAE   &    17.81 &	   15.01 &	  14.07	&    13.61 &    15.23 &    3.83	&    1.72 &    1.05 &    0.80 &    1.93 \\
CVAE+S &    18.43 &	   14.57 &\tb{13.18}&\tb{12.30}&    14.77 &    3.62	&    1.75 &    1.19 &    0.97 &    1.95 \\
\fvae  &\tb{15.93}&\tb{14.17}&    13.70 &	 13.48 &\tb{14.39}&\tb{3.14}&\tb{0.99}&\tb{0.74}&\tb{0.65}&\tb{1.45}\\ 
     \bottomrule
\end{tabular}
    \caption{Results on \fmnist. The best results are printed in bold.} 
	\label{tab_fmnist}
\end{table}

\begin{table}[h]
	\centering
	\tabcolsep=0.1cm
\begin{tabular}{l|ccccc|ccccc}
	\toprule
       &    \multicolumn{5}{c}{NLL in $10^{-2}$ BPD}          &     \multicolumn{5}{|c}{\msemin in $10^{-2}$}   \\ 
       &    0     &    1     &    2     &    3     &    avg   &    0    &    1    &    2    &    3    &    avg  \\
     \midrule
FCN    &          &          &          &          &          &   13.77 &   14.82 &   13.10 &   11.24 &   13.23 \\
FCN+S  &          &          &          &          &          &   12.56 &    8.96 &    6.06 &    4.09 &    7.92 \\
CVAE   &    446.0 &    280.1 &    273.5 &    266.5 &    316.7 &    9.23 &    3.46 &    2.27 &    1.55 &    4.14 \\
CVAE+S &    525.0 &    270.1 &    233.5 &\tb{203.5}&    308.3 &   11.08 &    5.49 &    3.66 &    2.57 &    5.71 \\
\fvae  &\tb{248.1}&\tb{227.6}&\tb{231.2}&    228.7 &\tb{233.9}&\tb{5.11}&\tb{0.88}&\tb{0.86}&\tb{0.84}&\tb{1.93}\\  
     \bottomrule
\end{tabular}
    \caption{Results on \fceleba. The best results are printed in bold.} 
	\label{tab_fceleba}
\end{table}

\begin{table}[h]
	\centering
	\tabcolsep=0.1cm
\begin{tabular}{l|ccccc|ccccc}
	\toprule
       &      \multicolumn{5}{c}{NLL in $10^{-2}$ BPD}        &     \multicolumn{5}{|c}{\msemin in $10^{-2}$}   \\  
       &    0     &    1     &    2     &    3     &    avg   &    0    &    1    &    2    &    3    &    avg  \\
     \midrule
FCN    &          &          &          &          &          &    5.83 &    3.34 &    2.37 &    1.82 &    3.35 \\
FCN+S  &          &          &          &          &          &    8.06 &    1.84 &    1.13 &    0.74 &    2.97 \\
CVAE   &    25.24 &    23.73 &    22.70 &    23.13 &    23.71 &    5.57 &    1.54 &    0.77 &    0.37 &    2.08 \\
CVAE+S &    26.08 &    24.94 &    23.98 &    23.95 &    24.75 &    4.95 &    2.50 &    1.77 &    1.19 &    2.62 \\
\fvae  &\tb{24.18}&\tb{23.07}&\tb{22.23}&\tb{22.88}&\tb{23.10}&\tb{4.11}&\tb{0.59}&\tb{0.32}&\tb{0.19}&\tb{1.32}\\  
     \bottomrule
\end{tabular}
    \caption{Results on \ftless. The best results are printed in bold.} 
	\label{tab_ftless}
\end{table}

\subsection{Qualitative Results}

\begin{figure}[h]
  \centering
  \begin{minipage}{0.9\textwidth}
  \centering
\begin{tikzpicture}
    \tikzset{every node/.append style={midway, font=\scriptsize, text depth=2.1ex}}
    \newcommand\gs{1}
    \newcommand\gl{16}
    \newcommand\res{32}
    \newcommand\Ni{15}
    \newcommand\tp{(\Ni * \res + 8 * \gs + 6 * \gl)}
    \newcommand\hy{0.06}
    \newcommand\lw{0.3mm} 
\draw [line width=\lw,  black] (0, \hy) -- ({((3*\res + 2*\gs)/\tp)*\columnwidth}, \hy) node {Input};
\draw [line width=\lw,  black] ({((3*\res + 2*\gs + 1*\gl)/\tp)*\columnwidth}, \hy) -- ({((4*\res + 2*\gs + 1*\gl)/\tp)*\columnwidth}, \hy) node {Target};
\draw [line width=\lw,  black] ({((4*\res + 2*\gs + 2*\gl)/\tp)*\columnwidth}, \hy) -- ({((7*\res + 4*\gs + 2*\gl)/\tp)*\columnwidth}, \hy) node {\fvae};
\draw [line width=\lw,  black] ({((7*\res + 4*\gs + 3*\gl)/\tp)*\columnwidth}, \hy) -- ({((10*\res + 6*\gs + 3*\gl)/\tp)*\columnwidth}, \hy) node {CVAE+S};
\draw [line width=\lw,  black] ({((10*\res + 6*\gs + 4*\gl)/\tp)*\columnwidth}, \hy) -- ({((13*\res + 8*\gs + 4*\gl)/\tp)*\columnwidth}, \hy) node {CVAE};
\draw [line width=\lw,  black] ({((13*\res + 8*\gs + 5*\gl)/\tp)*\columnwidth}, \hy) -- ({((14*\res + 8*\gs + 5*\gl)/\tp)*\columnwidth}, \hy) node {FCN+S};
\draw [line width=\lw,  black] ({((14*\res + 8*\gs + 6*\gl)/\tp)*\columnwidth}, \hy) -- ({((15*\res + 8*\gs + 6*\gl)/\tp)*\columnwidth}, \hy) node {FCN};
    
    \begin{scope}
        \node[anchor=north west, inner sep=0] (image) at (0,0) {\includegraphics[width=1.0\columnwidth]{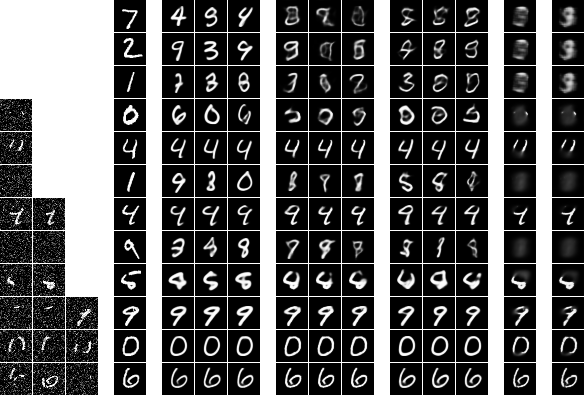}};
    \end{scope}
    \end{tikzpicture}%
    \end{minipage}%
    \caption{Prediction results of the different architectures on \fmnist for zero to three input images.}
    \label{fig_mnist_results}
\end{figure}

\begin{figure}[h]
  \centering
  \begin{minipage}{0.9\textwidth}
  \centering
\begin{tikzpicture}
    \tikzset{every node/.append style={midway, font=\scriptsize, text depth=2.1ex}}
    \newcommand\gs{2}
    \newcommand\gl{32}
    \newcommand\res{64}
    \newcommand\Ni{15}
    \newcommand\tp{(\Ni * \res + 8 * \gs + 6 * \gl)}
    \newcommand\hy{0.06} 
    \newcommand\lw{0.3mm} 
\draw [line width=\lw,  black] (0, \hy) -- ({((3*\res + 2*\gs)/\tp)*\columnwidth}, \hy) node {Input};
\draw [line width=\lw,  black] ({((3*\res + 2*\gs + 1*\gl)/\tp)*\columnwidth}, \hy) -- ({((4*\res + 2*\gs + 1*\gl)/\tp)*\columnwidth}, \hy) node {Target};
\draw [line width=\lw,  black] ({((4*\res + 2*\gs + 2*\gl)/\tp)*\columnwidth}, \hy) -- ({((7*\res + 4*\gs + 2*\gl)/\tp)*\columnwidth}, \hy) node {\fvae};
\draw [line width=\lw,  black] ({((7*\res + 4*\gs + 3*\gl)/\tp)*\columnwidth}, \hy) -- ({((10*\res + 6*\gs + 3*\gl)/\tp)*\columnwidth}, \hy) node {CVAE+S};
\draw [line width=\lw,  black] ({((10*\res + 6*\gs + 4*\gl)/\tp)*\columnwidth}, \hy) -- ({((13*\res + 8*\gs + 4*\gl)/\tp)*\columnwidth}, \hy) node {CVAE};
\draw [line width=\lw,  black] ({((13*\res + 8*\gs + 5*\gl)/\tp)*\columnwidth}, \hy) -- ({((14*\res + 8*\gs + 5*\gl)/\tp)*\columnwidth}, \hy) node {FCN+S};
\draw [line width=\lw,  black] ({((14*\res + 8*\gs + 6*\gl)/\tp)*\columnwidth}, \hy) -- ({((15*\res + 8*\gs + 6*\gl)/\tp)*\columnwidth}, \hy) node {FCN};
    
    \begin{scope}
        \node[anchor=north west, inner sep=0] (image) at (0,0) {\includegraphics[width=1.0\columnwidth]{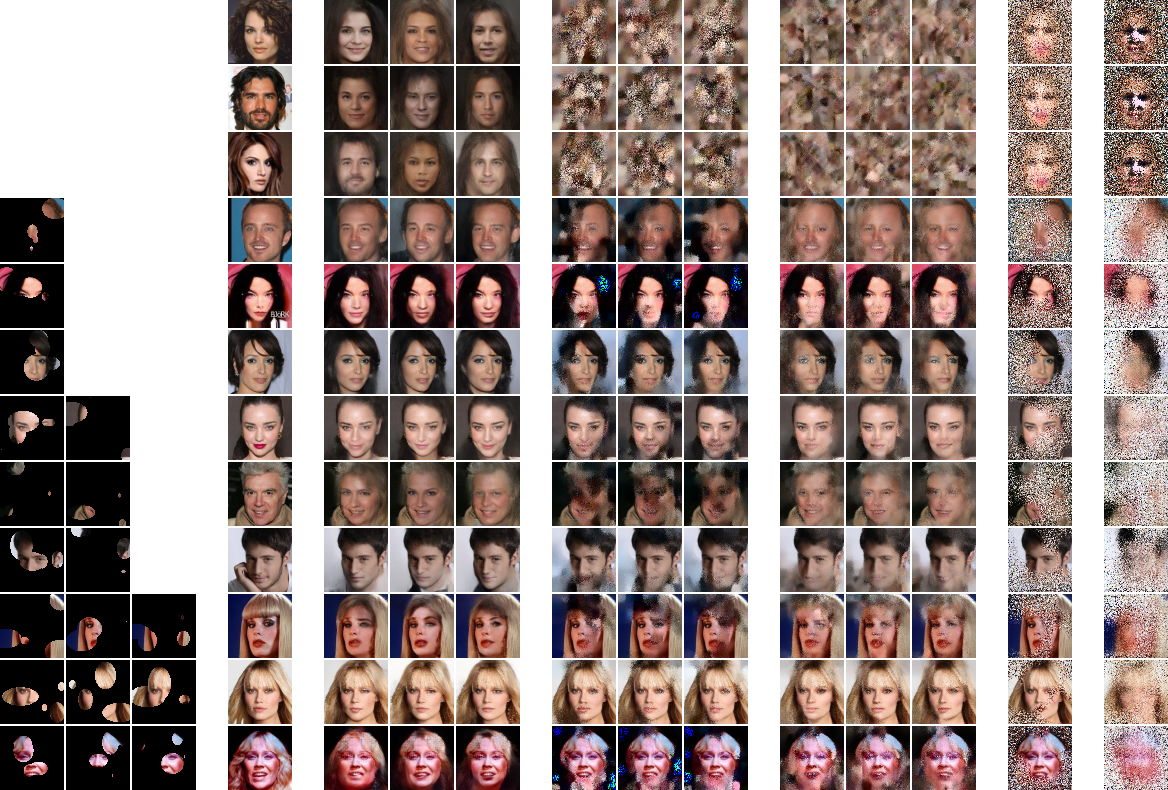}};
    \end{scope}
    \end{tikzpicture}%
    \end{minipage}
    \caption{Predictions on \fceleba for zero to three input images.}
    \label{fig_celeba_results}
\end{figure}

\begin{figure}[h]
  \centering 
  \begin{minipage}{0.9\textwidth}
  \centering
\begin{tikzpicture}
    \tikzset{every node/.append style={midway, font=\scriptsize, text depth=2.1ex}}
    \newcommand\gs{2}
    \newcommand\gl{32}
    \newcommand\res{64} 
    \newcommand\Ni{15}
    \newcommand\tp{(\Ni * \res + 8 * \gs + 6 * \gl)}
    \newcommand\hy{0.06}
    \newcommand\lw{0.3mm}
\draw [line width=\lw,  black] (0, \hy) -- ({((3*\res + 2*\gs)/\tp)*\columnwidth}, \hy) node {Input};
\draw [line width=\lw,  black] ({((3*\res + 2*\gs + 1*\gl)/\tp)*\columnwidth}, \hy) -- ({((4*\res + 2*\gs + 1*\gl)/\tp)*\columnwidth}, \hy) node {Target};
\draw [line width=\lw,  black] ({((4*\res + 2*\gs + 2*\gl)/\tp)*\columnwidth}, \hy) -- ({((7*\res + 4*\gs + 2*\gl)/\tp)*\columnwidth}, \hy) node {\fvae};
\draw [line width=\lw,  black] ({((7*\res + 4*\gs + 3*\gl)/\tp)*\columnwidth}, \hy) -- ({((10*\res + 6*\gs + 3*\gl)/\tp)*\columnwidth}, \hy) node {CVAE+S};
\draw [line width=\lw,  black] ({((10*\res + 6*\gs + 4*\gl)/\tp)*\columnwidth}, \hy) -- ({((13*\res + 8*\gs + 4*\gl)/\tp)*\columnwidth}, \hy) node {CVAE};
\draw [line width=\lw,  black] ({((13*\res + 8*\gs + 5*\gl)/\tp)*\columnwidth}, \hy) -- ({((14*\res + 8*\gs + 5*\gl)/\tp)*\columnwidth}, \hy) node {FCN+S};
\draw [line width=\lw,  black] ({((14*\res + 8*\gs + 6*\gl)/\tp)*\columnwidth}, \hy) -- ({((15*\res + 8*\gs + 6*\gl)/\tp)*\columnwidth}, \hy) node {FCN};

    \begin{scope}
        \node[anchor=north west, inner sep=0] (image) at (0,0) {\includegraphics[width=1.0\columnwidth]{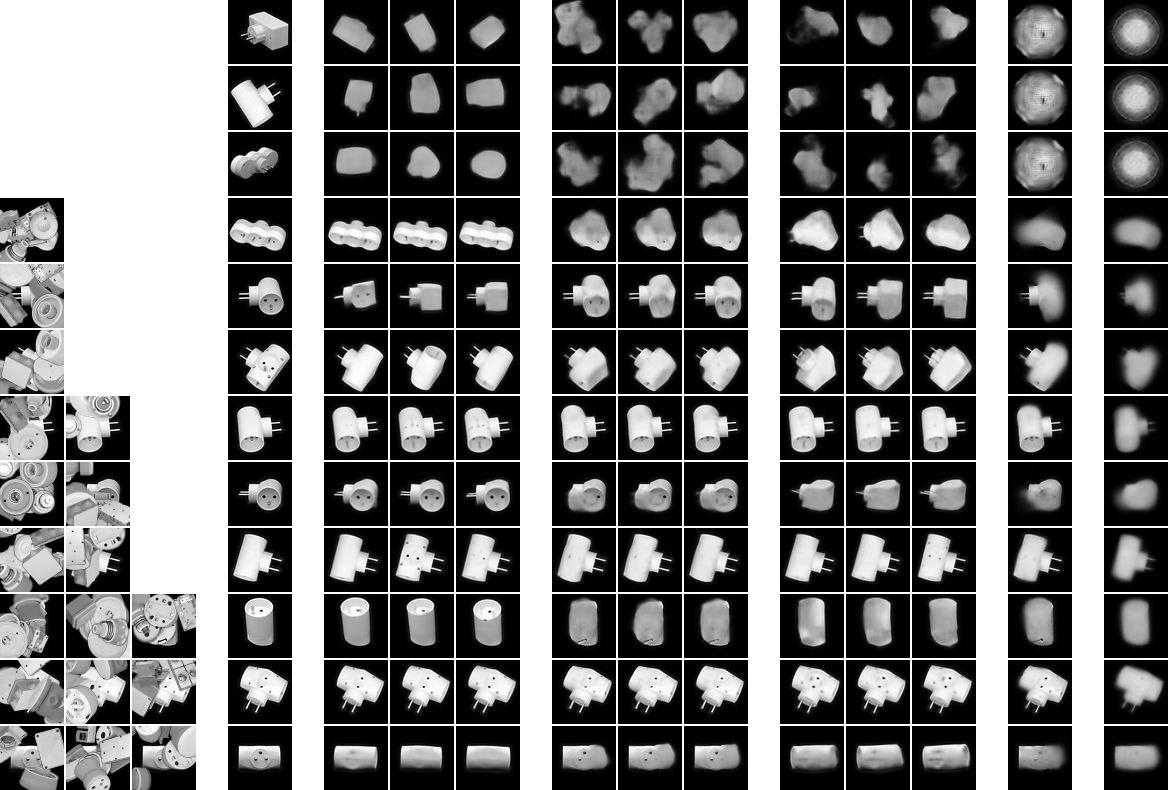}};
    \end{scope}
    \end{tikzpicture}%
    \end{minipage}
    \caption{Predictions on \ftless for zero to three input images.}
    \label{fig_tless_results}
\end{figure}
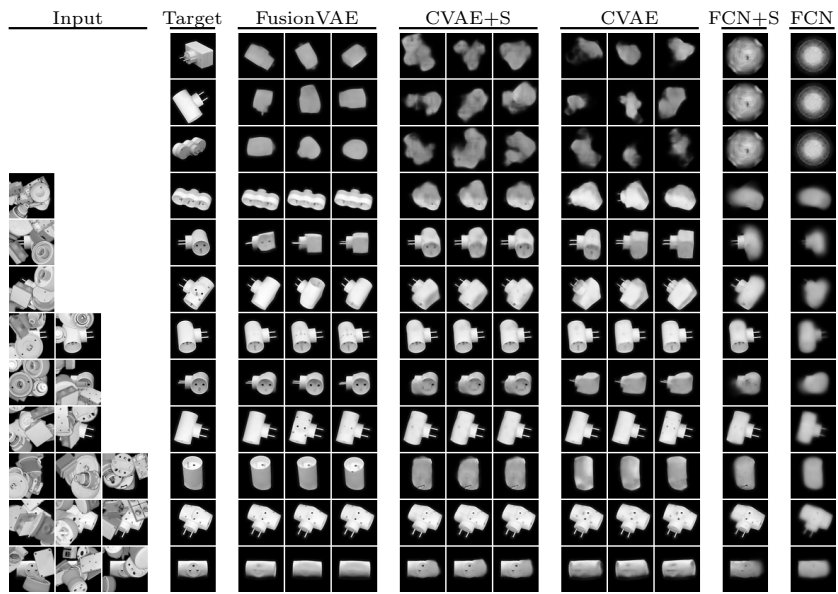

The outstanding performance of our architecture in comparison to the others is also obvious when looking at the qualitative results in \cref{fig_mnist_results,fig_celeba_results,fig_tless_results}. For every row, these figures show the input, target, and up to three output predictions for all architectures. For the FCN, we depict just a single output prediction per row as all of them look almost identical. 

In the first three rows when the network does not receive any input image, we see that our network provides very realistic images. This indicates that it is able to capture the underlying distribution of the used datasets very well and much better than the other architectures. Due to the difficulty of the \ftless dataset,
none of the models is able to produce realistic images without any input. Still our model shows much better performance in generating object-like shapes. In case the models receive at least one input image (cf. rows 4 -- 12), all architectures are able to extract the available information from the given input images. In addition, all VAE approaches, ours included, are able to complete the given input data based on prior knowledge. 
It is clearly visible, however, that the predictions of our model are much more realistic than the ones of the standard CVAE approaches especially for the more difficult datasets like \fceleba and \ftless.

\clearpage

\subsection{Ablation Studies}
\label{sec_ablation}

We conducted ablation studies to show the effect of certain design choices, such as the selection of the approximate posterior and the aggregation method. All experiments are run on the \fceleba dataset.

\cref{tab_posterior_ablation} compares the performance of our \fvae for two different approximate posterior distributions $q$. The approximate posterior we selected for our \fvae $q(y)$, depends only on the given target image $y$. It performs slightly better on average compared to the same method using a posterior that is computed based on the input images $x$ as well as the target image $y$. However, the latter approach is superior when fusing two or three input images.

\begin{table}[h]
    \centering
    \begin{tabular}{l|ccccc|ccccc}
    \toprule
      & \multicolumn{5}{c}{NLL in $10^{-2}$ BPD} & \multicolumn{5}{|c}{\msemin in $10^{-2}$} \\
      & \multicolumn{1}{c}{0} & \multicolumn{1}{c}{1} & \multicolumn{1}{c}{2} & \multicolumn{1}{c}{3} & \multicolumn{1}{c}{avg} & \multicolumn{1}{|c}{0} & \multicolumn{1}{c}{1} & \multicolumn{1}{c}{2} & \multicolumn{1}{c}{3} & \multicolumn{1}{c}{avg} \\
    \midrule
    		                   				
$q(x,y)$ &    309.2	&    246.8 &\tb{211.1}&\tb{180.5}&    237.0 &\tb{5.04}&    1.50 &    0.95 &\tb{0.66}&    2.04 \\
$q(y)$   &\tb{248.1}&\tb{227.6}&    231.2 &    228.7 &\tb{233.9}&    5.11 &\tb{0.88}&\tb{0.86}&    0.84 &\tb{1.93}\\  
    \bottomrule
    \end{tabular}
    \caption{Posterior ablation on the \fceleba dataset. The best results are printed in bold.}
	\label{tab_posterior_ablation}
\end{table}

\begin{table}[h]

    \centering
    \begin{tabular}{l|rrrrr|ccccc}
    \toprule
      & \multicolumn{5}{c}{NLL in $10^{-2}$ BPD} & \multicolumn{5}{|c}{\msemin in $10^{-2}$} \\
      & \multicolumn{1}{c}{0} & \multicolumn{1}{c}{1} & \multicolumn{1}{c}{2} & \multicolumn{1}{c}{3} & \multicolumn{1}{c}{avg} & \multicolumn{1}{|c}{0} & \multicolumn{1}{c}{1} & \multicolumn{1}{c}{2} & \multicolumn{1}{c}{3} & \multicolumn{1}{c}{avg} \\
    \midrule
MaxAggAdd  &\tb{248.1}&    227.6 &    231.2 &    228.7 &    233.9 &    5.11 &\tb{0.88}&    0.86 &    0.84 &\tb{1.93}\\ 
MeanAggAdd &    270.9 &\tb{223.3}&\tb{216.4}&    214.4 &    231.3 &    5.41 &    1.00 &\tb{0.79}&\tb{0.70}&    1.98 \\
BayAggAdd  &    970.7 &    294.0 &    291.4 &    291.5 &    462.6 &    6.03 &    5.17 &    5.10 &    5.12 &    5.36 \\
MaxAggAll  &    249.6 &    236.0 &    223.9 &\tb{212.7}&\tb{230.6}&    6.15 &    2.82 &    1.84 &    1.30 &    3.03 \\
MeanAggAll &    252.7 &    235.2 &    222.2 &    213.5 &    230.9 &    6.19 &    2.39 &    1.52 &    1.13 &    2.81 \\
BayAggAll  &    255.6 &    568.3 &    414.7 &   1376.9 &    653.3 &\tb{5.10}&    4.24 &    1.96 &    1.39 &    3.18 \\
    \bottomrule
    \end{tabular}
    \caption{Aggregation ablation on the \fceleba dataset. The best results are printed in bold.}
	\label{tab_aggregation_ablation}
\end{table}

\cref{tab_aggregation_ablation} shows the performance of different aggregation methods which are applied to create the prior distributions $p_l$ of every latent group. In our \fvae, the prior is created by fusing the input image features $f_{lx}$ using max aggregation (MaxAgg) and adding them to the decoded features of the same latent group before applying a 2D convolution. We abbreviate that method with MaxAggAdd. 

In addition to MaxAgg, we examined mean aggregation (MeanAgg) and Bayesian aggregation (BayAgg) \cite{baco} for comparison.
For each aggregation principle, we tried two different versions: 1) aggregation of the input image features $\boldsymbol f_{lx}$ adding the corresponding information from the decoder in a pixel-wise manner (denoted by suffix Add), and 2) directly aggregating all features, i.e. both input image features $\boldsymbol f_{lx}$ and decoder features (denoted by suffix All).

For creating the prior $p_i$ when using BayAgg, we moved the 2D convolutions before the aggregation in order to create the parameters $\mu$ and $\sigma$ of a latent Gaussian distribution. Unlike MaxAgg and MeanAgg, BayAgg directly outputs a new Gaussian distribution that does not need to be processed any further by a convolution.

We can see that all variations of mean and max aggregation are significantly better than Bayesian aggregation. Also their training procedures are less often impaired due to numeric instabilities. Interestingly, the NLL is very similar independent of whether the aggregation is performed on all features or not. However, the \msemin is much better for the aggregation with addition. Since the expressiveness of the metrics is limited, we provide additional visualizations of this ablation in \cref{app_sec_ablation}.

\section{Conclusion}

We have presented a novel deep hierarchical variational autoencoder for generative image fusion called \fvae. Our approach fuses multiple corrupted input images together with prior knowledge obtained during training. 
We created three challenging image fusion benchmarks based on common computer vision datasets. Moreover, we implemented four standard methods that we modified to support our tasks. We showed that our \fvae outperforms all other methods significantly while having a similar number of trainable parameters. The predicted images of our approach look very realistic and incorporate given input information almost perfectly. During ablation studies, we revealed the benefits of our design choices regarding the applied aggregation method and the used posterior distribution. In future work, our research could be extended by enabling the fusion of different modalities e.g. by using multiple encoders. Additionally, an explicit uncertainty estimation could be implemented that helps to weigh the impact of input information according to its uncertainty.


\clearpage
%
%
\bibliographystyle{splncs04}
\bibliography{egbib}


\appendix

\ifsupplementaryheader
\clearpage
{\centering
\Large

\textbf{ Supplementary Material}

}
\appendix
\fi

\section{Implementation details}

For \fmnist and \ftless, we model the decoder's output by pixel-wise independent Bernoulli distributions. For \fceleba, we use pixel-wise independent discretized logistic mixture distributions as proposed by Salimans et al. \cite{pixelcnnpp}.

The residual cells of the encoder are composed of batch normalization layers \cite{batchNorm}, Swish activation functions \cite{swish}, convolutional layers, and Squeeze-and-Excitation (SE) blocks \cite{squeezeExcitation} as proposed in \cite{nvae}.
In the decoder, we also follow \cite{nvae} and build the residual cells out of batch normalization layers, 1x1 convolutions, Swish activations, depthwise separable convolutions \cite{depSepConv}, and SE blocks.
However, we omitted normalizing flow because in our experiments it showed to increase the training time without improving the prediction accuracy significantly. 

For each dataset, we chose the size of the architecture individually to achieve acceptable accuracy while keeping the training time reasonable. 
\cref{tab_implementation_details} provides details about the used hyperparameters.

\begin{table*}[h]
    \centering
    \begin{tabular}{lccc}
    \toprule
 Hyperparameter & \fmnist & \fceleba & \ftless \\
    \midrule
 \# latent groups per scale  & 5, 2 & 10, 5, 2 & 10, 5, 2 \\
 spatial dimensions of $\boldsymbol z_l$ per scale  & $4^2, 8^2$ & $8^2, 16^2, 32^2$ & $8^2, 16^2, 32^2$ \\
 \# channels in $\boldsymbol z_l$ & 10 & 20 & 20 \\
 \# GPUs & 2 & 4 & 2 \\
 \# training epochs & 400 & 90 & 500 \\
  batch size & 800 & 32 & 32 \\
 Training time & 4h & 48h & 28h \\
    \bottomrule
    \end{tabular}
    \caption{Main hyperparameters of our experiments.}
	\label{tab_implementation_details}
\end{table*}

In general, the number of latent groups $L$ should be chosen depending on the complexity of the task at hand. We made our decision based on the $L$ of the NVAE \cite{nvae} but reduced it for computational reasons. For \fceleba and \ftless, we use 17 latent groups, for \fmnist only seven. Using more latent groups improves the results but increases the computational effort significantly.

For all experiments, we used GPUs of type NVIDIA Tesla V100 with 32GB of memory and trained with an AdaMax optimizer \cite{adam}. We applied a cosine annealing schedule for the learning rate \cite{sgdr} starting at 0.01 and ending at 0.0001.

\clearpage

\section{Derivation of Bayesian Aggregation}
\label{sec_derivation_of_bayesian_aggregation}

We use two related encoders to learn a latent observation $\bbs\mu_i=\text{enc}_\mu(\bbx_i, \bby)$ with its corresponding variance values $\bbs\sigma_i=\text{enc}_\sigma(\bbx_i, \bby)$.

Assuming a factorized Gaussian prior distribution in the latent space 
$p(\bbz) = \mathcal{N}(\bbz|\bbs\mu_{\bbz, 0}, \text{diag}(\bbs\sigma_{\bbz, 0}))$,
we can derive the factorized posterior distribution
$q_\phi(\bbz|\bby) = \mathcal{N}(\bbz|\bbs\mu_z, \text{diag}(\bbs\sigma_z)$
in closed form using standard Gaussian conditioning \cite{bishop2006} following \cite{baco}
\begin{equation}
    \bbs\sigma_{\bbz}^2 
    = \left[ (\bbs\sigma_{\bbz, 0}^2)^\ominus + (\bbs\sigma_{i}^2)^\ominus \right]^\ominus,
\end{equation}
\begin{equation}
    \bbs\mu_{\bbz} = 
    \bbs\mu_{\bbz, 0} + \bbs\sigma_{\bbz, 0}^2 \odot (\bbs\mu_i - \bbs\mu_{\bbz, 0}) \oslash \bbs\sigma_{i}^2
\end{equation}
where $\ominus$ denotes element-wise inversion, $\odot$ denotes element-wise multiplication, and $\oslash$ denotes element-wise division.

\section{Derivation of \fvae's ELBO}
\label{elbo_derivation}

We start with the following KL divergence between the approximate posterior and the real posterior,
\begin{equation}
\mathrm{KL} (q_{\theta}(\bbz|\bby) || p_{\theta}(\bbz|\bbx,\bby)) \ge 0.
\end{equation}
Next, we apply the Bayes's theorem to obtain
\begin{equation}
-\int q_{\theta}(\bbz|\bby) \log \frac{p(\bby|\bbx,\bbz) p(\bbz|\bbx)}{p(\bby|\bbx)q_{\theta}(\bbz|\bby)} dz\ge 0.
\end{equation}
This leads to
\begin{align}
\begin{split}
-\mathbb{E}_{q_{\theta}(\bbz|\bby)} [\log p(\bby|\bbx,\bbz)]  & - \mathrm{KL} (q_{\theta} (\bbz|\bby)|| p(\bbz|\bbx))  \\
&+ \int q_{\theta}(\bbz|\bby) \log p(\bby|\bbx)  dz \ge 0.
\end{split}
\end{align}
The term $\log p(\bby|\bbx)$ can be moved out from the third integral component, and leaves the integral becoming 1. Finally, we obtain the ELBO of the conditional log-likelihood
\begin{equation}
\log p(\bby|\bbx) \ge \mathbb{E}_{q_{\theta}(\bbz|\bby)} [\log p(\bby|\bbx,\bbz)]  + \mathrm{KL} (q_{\theta} (\bbz|\bby)|| p(\bbz|\bbx))  .
\end{equation}

\clearpage

\section{Ablation Studies}
\label{app_sec_ablation}

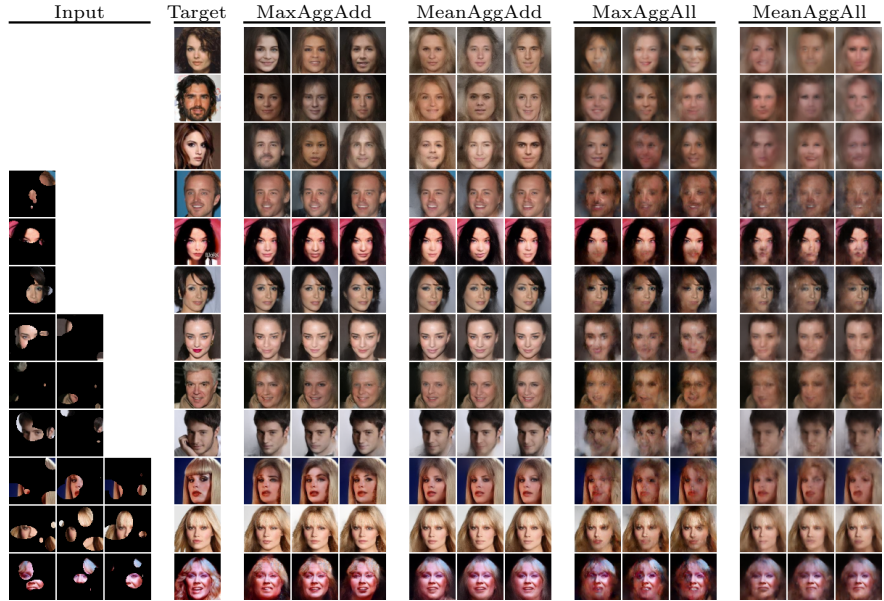
\begin{figure}
  \centering
  \begin{minipage}{0.95\textwidth}
  \centering
\begin{tikzpicture}
\tikzset{every node/.append style={midway, font=\scriptsize, text depth=2.1ex}}
\newcommand\gs{2}
\newcommand\gl{32}
\newcommand\res{64}
\newcommand\Ni{16}
\newcommand\tp{(\Ni * \res + 10 * \gs + 5 * \gl)}
\newcommand\hy{0.06}
\newcommand\lw{0.3mm}
\draw [line width=\lw,  black] (0, \hy) -- ({((3*\res + 2*\gs)/\tp)*\columnwidth}, \hy) node {Input};
\draw [line width=\lw,  black] ({((3*\res + 2*\gs + 1*\gl)/\tp)*\columnwidth}, \hy) -- ({((4*\res + 2*\gs + 1*\gl)/\tp)*\columnwidth}, \hy) node {Target};
\draw [line width=\lw,  black] ({((4*\res + 2*\gs + 2*\gl)/\tp)*\columnwidth}, \hy) -- ({((7*\res + 4*\gs + 2*\gl)/\tp)*\columnwidth}, \hy) node {MaxAggAdd};
\draw [line width=\lw,  black] ({((7*\res + 4*\gs + 3*\gl)/\tp)*\columnwidth}, \hy) -- ({((10*\res + 6*\gs + 3*\gl)/\tp)*\columnwidth}, \hy) node {MeanAggAdd};
\draw [line width=\lw,  black] ({((10*\res + 6*\gs + 4*\gl)/\tp)*\columnwidth}, \hy) -- ({((13*\res + 8*\gs + 4*\gl)/\tp)*\columnwidth}, \hy) node {MaxAggAll};
\draw [line width=\lw,  black] ({((13*\res + 8*\gs + 5*\gl)/\tp)*\columnwidth}, \hy) -- ({((16*\res + 10*\gs + 5*\gl)/\tp)*\columnwidth}, \hy) node {MeanAggAll};

\begin{scope}
    \node[anchor=north west, inner sep=0] (image) at (0,0) {\includegraphics[width=1.0\columnwidth]{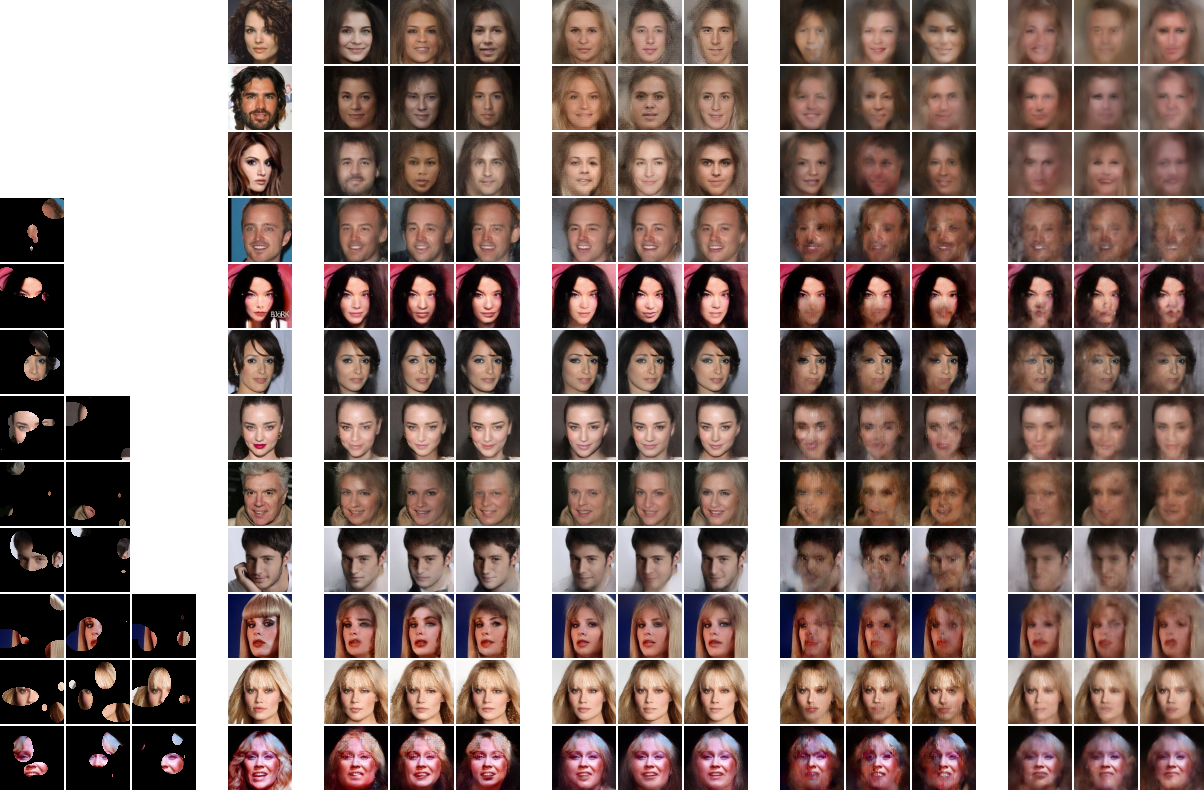}};
\end{scope}
\end{tikzpicture}
\end{minipage}
 \caption{Prediction results of the different aggregation methods on \fceleba for zero to three input images.}
    \label{fig_agg_ablation}
\end{figure}

This is a supplement for the aggregation ablation study in \cref{sec_ablation}. In \cref{tab_aggregation_ablation}, we saw that the average NLL of all experiments using mean and max aggregation methods are similar. \cref{fig_agg_ablation} shows the corresponding qualitative results.
However, even though the NLL is very similar, the results of the aggregation of all features (MaxAggAll and MeanAggAll) are much more blurry than the results of the aggregation with addition (MaxAggAdd and MeanAggAdd). This is in conformity with the \msemin results. It indicates that the NLL alone is not always the best metric to assess the visual closeness to real faces. 
When carefully examining the images of the addition aggregations, you could argue that the predictions with zero input images look slightly more realistic for max aggregation while for three input images, mean aggregation seems to be marginally better. This again confirms the validity of the \msemin results even though the NLL results are also in accordance for this comparison.

\newpage
\section{Statistic Significance of the Results}

All experiments for this publications were carefully designed and optimized so that the training procedures are stable and lead to reproducible results. However, the data processing pipelines introduce randomness which lead to non-deterministic training outcomes due to multi-GPU training. We therefore ran every experiment three times and reported the results of the best training in \cref{sec:results}. In \cref{tab_fmnist_stat_nll,tab_fmnist_stat_mse,tab_fceleba_stat_nll,tab_fceleba_stat_mse,tab_ftless_stat_nll,tab_ftless_stat_mse} we provide the means and variances of the three training runs.

\begin{table}[h]
	\centering
	\small
\begin{tabular}{l|rrrrr}
	\toprule
  & \multicolumn{1}{c}{0} & \multicolumn{1}{c}{1} & \multicolumn{1}{c}{2} & \multicolumn{1}{c}{3} & \multicolumn{1}{c}{avg} \\
     \midrule                        
CVAE   & $    17.67 \pm0.10$ & $    15.11 \pm0.07$ & $    14.19 \pm0.08$ & $    13.71 \pm0.07$ & $    15.27 \pm0.02$ \\  
CVAE+S & $    18.45 \pm0.02$ & $    14.64 \pm0.06$ & $\tb{13.22}\pm0.03$ & $\tb{12.32}\pm0.02$ & $    14.81 \pm0.03$ \\  
\fvae  & $\tb{15.91}\pm0.03$ & $\tb{14.13}\pm0.07$ & $    13.64 \pm0.09$ & $    13.41 \pm0.10$ & $\tb{14.34}\pm0.07$ \\  
     \bottomrule
\end{tabular}
    \caption{Mean and standard deviation of the \fmnist NLL results in $10^{-2}$ BPD. The best results are printed in bold.} 
	\label{tab_fmnist_stat_nll}
\end{table}

\begin{table}[h]
	\centering
	\small
\begin{tabular}{l|rrrrr}
	\toprule
  & \multicolumn{1}{c}{0} & \multicolumn{1}{c}{1} & \multicolumn{1}{c}{2} & \multicolumn{1}{c}{3} & \multicolumn{1}{c}{avg} \\
     \midrule                        
FCN    & $   10.84 \pm0.37$ & $    5.96 \pm0.11$ & $    6.02 \pm0.18$ & $    6.13 \pm0.25$ & $    7.38 \pm 0.11$ \\  
FCN+S  & $    6.21 \pm0.65$ & $    3.79 \pm0.04$ & $    2.64 \pm0.07$ & $    1.88 \pm0.08$ & $    3.73 \pm 0.22$ \\  
CVAE   & $    3.87 \pm0.03$ & $    1.76 \pm0.03$ & $    1.09 \pm0.03$ & $    0.83 \pm0.02$ & $    1.97 \pm 0.03$ \\  
CVAE+S & $    3.53 \pm0.06$ & $    1.77 \pm0.01$ & $    1.23 \pm0.04$ & $    1.02 \pm0.04$ & $    1.96 \pm 0.01$ \\  
\fvae  & $\tb{3.14}\pm0.01$ & $\tb{1.04}\pm0.06$ & $\tb{0.77}\pm0.04$ & $\tb{0.67}\pm0.03$ & $\tb{1.47}\pm0.03$ \\  
     \bottomrule
\end{tabular}
    \caption{Mean and standard deviation of the \fmnist \msemin results in $10^{-2}$. The best results are printed in bold.} 
	\label{tab_fmnist_stat_mse}
\end{table}

\begin{table}[h]
	\centering
	\small
\begin{tabular}{l|rrrrr}
	\toprule
  & \multicolumn{1}{c}{0} & \multicolumn{1}{c}{1} & \multicolumn{1}{c}{2} & \multicolumn{1}{c}{3} & \multicolumn{1}{c}{avg} \\
     \midrule                        
CVAE   & $    456.9 \pm~9.42$ & $    289.1 \pm~6.53$ & $    278.9 \pm~4.51$ & $    270.3 \pm~3.66$ & $    324.0 \pm~5.17$ \\  
CVAE+S & $    487.4 \pm27.45$ & $    355.4 \pm60.39$ & $    280.7 \pm35.12$ & $    230.9 \pm22.59$ & $    338.8 \pm22.99$ \\  
\fvae  & $\tb{251.0}\pm~2.06$ & $\tb{222.3}\pm~3.71$ & $\tb{226.8}\pm~3.16$ & $\tb{224.0}\pm~3.36$ & $\tb{231.0}\pm~2.05$ \\  
     \bottomrule
\end{tabular}
    \caption{Mean and standard deviation of the \fceleba NLL results in $10^{-2}$ BPD. The best results are printed in bold.} 
	\label{tab_fceleba_stat_nll}
\end{table}

\begin{table}[h]
	\centering
	\small
\begin{tabular}{l|rrrrr}
	\toprule
  & \multicolumn{1}{c}{0} & \multicolumn{1}{c}{1} & \multicolumn{1}{c}{2} & \multicolumn{1}{c}{3} & \multicolumn{1}{c}{avg} \\
     \midrule                        
FCN    & $    13.07 \pm0.87$ & $    20.00 \pm3.77$ & $    17.87 \pm3.42$ & $    15.56 \pm3.08$ & $    16.62 \pm2.48$ \\  
FCN+S  & $    11.94 \pm0.52$ & $    18.58 \pm7.02$ & $    14.90 \pm6.59$ & $    11.19 \pm5.39$ & $    14.15 \pm4.65$ \\  
CVAE   & $     8.70 \pm0.42$ & $     6.25 \pm2.16$ & $     4.33 \pm1.77$ & $     3.02 \pm1.37$ & $     5.58 \pm1.21$ \\  
CVAE+S & $     9.87 \pm1.10$ & $     9.60 \pm3.34$ & $     7.30 \pm3.07$ & $     5.57 \pm2.64$ & $     8.09 \pm2.19$ \\  
\fvae  & $\tb{ 5.82}\pm0.52$ & $\tb{ 1.10}\pm0.16$ & $\tb{ 0.93}\pm0.06$ & $\tb{ 0.84}\pm0.03$ & $\tb{ 2.18}\pm0.17$ \\  
     \bottomrule
\end{tabular}
    \caption{Mean and standard deviation of the \fceleba \msemin results in $10^{-2}$. The best results are printed in bold.} 
	\label{tab_fceleba_stat_mse}
\end{table}

\begin{table}[h]
	\centering
	\small
\begin{tabular}{l|rrrrr}
	\toprule
  & \multicolumn{1}{c}{0} & \multicolumn{1}{c}{1} & \multicolumn{1}{c}{2} & \multicolumn{1}{c}{3} & \multicolumn{1}{c}{avg} \\
     \midrule                        
CVAE   & $    25.27 \pm0.04$ & $    24.00 \pm0.25$ & $    22.98 \pm0.24$ & $    23.34 \pm0.19$ & $    23.90 \pm0.17$ \\  
CVAE+S & $    26.12 \pm0.23$ & $    25.29 \pm0.27$ & $    24.27 \pm0.25$ & $    24.15 \pm0.20$ & $    24.97 \pm0.16$ \\  
\fvae  & $\tb{24.32}\pm0.10$ & $\tb{23.09}\pm0.02$ & $\tb{22.25}\pm0.02$ & $\tb{22.90}\pm0.02$ & $\tb{23.15}\pm0.04$ \\  
     \bottomrule
\end{tabular}
    \caption{Mean and standard deviation of the \ftless NLL results in $10^{-2}$ BPD. The best results are printed in bold.} 
	\label{tab_ftless_stat_nll}
\end{table}

\begin{table}[h]
	\centering
	\small
\begin{tabular}{l|rrrrr}
	\toprule
  & \multicolumn{1}{c}{0} & \multicolumn{1}{c}{1} & \multicolumn{1}{c}{2} & \multicolumn{1}{c}{3} & \multicolumn{1}{c}{avg} \\
     \midrule                        
FCN    & $    5.88 \pm0.04$ & $    3.32 \pm0.10$ & $    2.50 \pm0.09$ & $    1.96 \pm0.10$ & $    3.43 \pm 0.06$ \\  
FCN+S  & $    8.83 \pm1.05$ & $    1.95 \pm0.14$ & $    1.28 \pm0.15$ & $    0.86 \pm0.14$ & $    3.26 \pm 0.37$ \\  
CVAE   & $    5.49 \pm0.11$ & $    1.73 \pm0.22$ & $    0.95 \pm0.18$ & $    0.44 \pm0.07$ & $    2.18 \pm 0.13$ \\  
CVAE+S & $    4.87 \pm0.06$ & $    2.98 \pm0.29$ & $    2.06 \pm0.19$ & $    1.27 \pm0.09$ & $    2.81 \pm 0.12$ \\  
\fvae  & $\tb{4.15}\pm0.03$ & $\tb{0.62}\pm0.03$ & $\tb{0.33}\pm0.03$ & $\tb{0.20}\pm0.02$ & $\tb{1.34}\pm 0.02$ \\  
     \bottomrule
\end{tabular}
    \caption{Mean and standard deviation of the \ftless \msemin results in $10^{-2}$. The best results are printed in bold.} 
	\label{tab_ftless_stat_mse}
\end{table}

\clearpage
\section{Reconstruction}

\begin{figure}[h]
  \centering
  \begin{minipage}{0.9\textwidth}
  \centering
\begin{tikzpicture}
    \tikzset{every node/.append style={midway, font=\scriptsize, text depth=2.1ex}}
    \newcommand\gs{1}
    \newcommand\gl{16} 
    \newcommand\res{32}
    \newcommand\Ni{15} 
    \newcommand\tp{(\Ni * \res + 8 * \gs + 6 * \gl)}  
    \newcommand\hy{0.06} 
    \newcommand\lw{0.3mm} 
\draw [line width=\lw,  black] (0, \hy) -- ({((3*\res + 2*\gs)/\tp)*\columnwidth}, \hy) node {Input};
\draw [line width=\lw,  black] ({((3*\res + 2*\gs + 1*\gl)/\tp)*\columnwidth}, \hy) -- ({((4*\res + 2*\gs + 1*\gl)/\tp)*\columnwidth}, \hy) node {Target};
\draw [line width=\lw,  black] ({((4*\res + 2*\gs + 2*\gl)/\tp)*\columnwidth}, \hy) -- ({((7*\res + 4*\gs + 2*\gl)/\tp)*\columnwidth}, \hy) node {\fvae};
\draw [line width=\lw,  black] ({((7*\res + 4*\gs + 3*\gl)/\tp)*\columnwidth}, \hy) -- ({((10*\res + 6*\gs + 3*\gl)/\tp)*\columnwidth}, \hy) node {CVAE+S};
\draw [line width=\lw,  black] ({((10*\res + 6*\gs + 4*\gl)/\tp)*\columnwidth}, \hy) -- ({((13*\res + 8*\gs + 4*\gl)/\tp)*\columnwidth}, \hy) node {CVAE};
\draw [line width=\lw,  black] ({((13*\res + 8*\gs + 5*\gl)/\tp)*\columnwidth}, \hy) -- ({((14*\res + 8*\gs + 5*\gl)/\tp)*\columnwidth}, \hy) node {FCN+S};
\draw [line width=\lw,  black] ({((14*\res + 8*\gs + 6*\gl)/\tp)*\columnwidth}, \hy) -- ({((15*\res + 8*\gs + 6*\gl)/\tp)*\columnwidth}, \hy) node {FCN};
    
    \begin{scope}
        \node[anchor=north west, inner sep=0] (image) at (0,0) {\includegraphics[width=1.0\columnwidth]{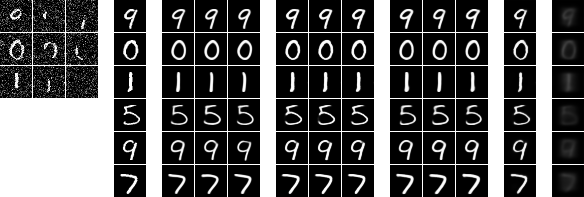}};
    \end{scope}
    \end{tikzpicture}
    \end{minipage}
    \caption{Reconstruction results of the different architectures on \fmnist.}
    \label{fig_fmnist_reconstruction}
\end{figure}
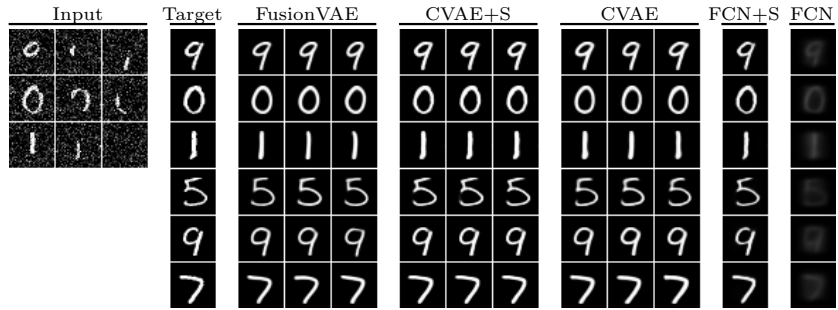

\begin{figure}[h]
  \centering
  \begin{minipage}{0.9\textwidth}
  \centering
\begin{tikzpicture}
    \tikzset{every node/.append style={midway, font=\scriptsize, text depth=2.1ex}}
    \newcommand\gs{2}
    \newcommand\gl{32}
    \newcommand\res{64}
    \newcommand\Ni{15} 
    \newcommand\tp{(\Ni * \res + 8 * \gs + 6 * \gl)} 
    \newcommand\hy{0.06} 
    \newcommand\lw{0.3mm}
\draw [line width=\lw,  black] (0, \hy) -- ({((3*\res + 2*\gs)/\tp)*\columnwidth}, \hy) node {Input};
\draw [line width=\lw,  black] ({((3*\res + 2*\gs + 1*\gl)/\tp)*\columnwidth}, \hy) -- ({((4*\res + 2*\gs + 1*\gl)/\tp)*\columnwidth}, \hy) node {Target};
\draw [line width=\lw,  black] ({((4*\res + 2*\gs + 2*\gl)/\tp)*\columnwidth}, \hy) -- ({((7*\res + 4*\gs + 2*\gl)/\tp)*\columnwidth}, \hy) node {\fvae};
\draw [line width=\lw,  black] ({((7*\res + 4*\gs + 3*\gl)/\tp)*\columnwidth}, \hy) -- ({((10*\res + 6*\gs + 3*\gl)/\tp)*\columnwidth}, \hy) node {CVAE+S};
\draw [line width=\lw,  black] ({((10*\res + 6*\gs + 4*\gl)/\tp)*\columnwidth}, \hy) -- ({((13*\res + 8*\gs + 4*\gl)/\tp)*\columnwidth}, \hy) node {CVAE};
\draw [line width=\lw,  black] ({((13*\res + 8*\gs + 5*\gl)/\tp)*\columnwidth}, \hy) -- ({((14*\res + 8*\gs + 5*\gl)/\tp)*\columnwidth}, \hy) node {FCN+S};
\draw [line width=\lw,  black] ({((14*\res + 8*\gs + 6*\gl)/\tp)*\columnwidth}, \hy) -- ({((15*\res + 8*\gs + 6*\gl)/\tp)*\columnwidth}, \hy) node {FCN};
    
    \begin{scope}
        \node[anchor=north west, inner sep=0] (image) at (0,0) {\includegraphics[width=1.0\columnwidth]{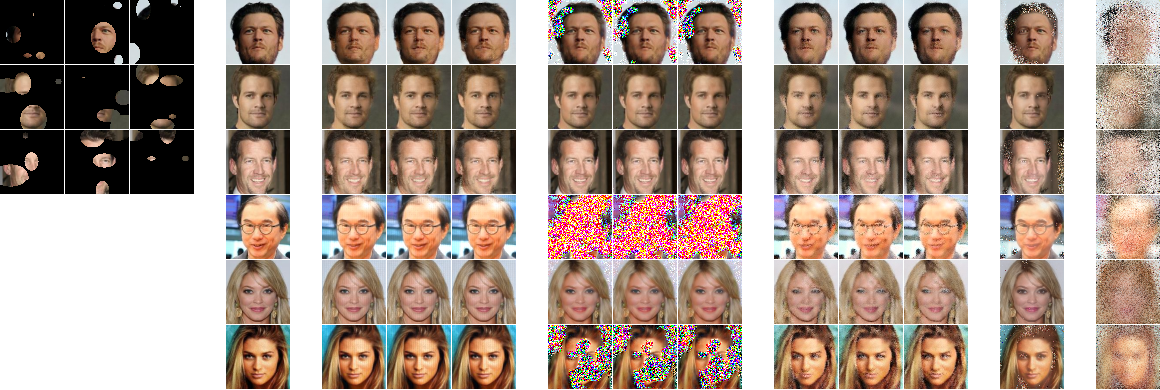}};
    \end{scope}
    \end{tikzpicture}
    \end{minipage}
    \caption{Reconstruction results on \fceleba.}
    \label{fig_fceleba_reconstruction}
\end{figure}

\begin{figure}[h]
  \centering
  \begin{minipage}{0.9\textwidth}
  \centering
\begin{tikzpicture}
    \tikzset{every node/.append style={midway, font=\scriptsize, text depth=2.1ex}}
    \newcommand\gs{2} 
    \newcommand\gl{32} 
    \newcommand\res{64}
    \newcommand\Ni{15}
    \newcommand\tp{(\Ni * \res + 8 * \gs + 6 * \gl)}  
    \newcommand\hy{0.06}
    \newcommand\lw{0.3mm}
\draw [line width=\lw,  black] (0, \hy) -- ({((3*\res + 2*\gs)/\tp)*\columnwidth}, \hy) node {Input};
\draw [line width=\lw,  black] ({((3*\res + 2*\gs + 1*\gl)/\tp)*\columnwidth}, \hy) -- ({((4*\res + 2*\gs + 1*\gl)/\tp)*\columnwidth}, \hy) node {Target};
\draw [line width=\lw,  black] ({((4*\res + 2*\gs + 2*\gl)/\tp)*\columnwidth}, \hy) -- ({((7*\res + 4*\gs + 2*\gl)/\tp)*\columnwidth}, \hy) node {\fvae};
\draw [line width=\lw,  black] ({((7*\res + 4*\gs + 3*\gl)/\tp)*\columnwidth}, \hy) -- ({((10*\res + 6*\gs + 3*\gl)/\tp)*\columnwidth}, \hy) node {CVAE+S};
\draw [line width=\lw,  black] ({((10*\res + 6*\gs + 4*\gl)/\tp)*\columnwidth}, \hy) -- ({((13*\res + 8*\gs + 4*\gl)/\tp)*\columnwidth}, \hy) node {CVAE};
\draw [line width=\lw,  black] ({((13*\res + 8*\gs + 5*\gl)/\tp)*\columnwidth}, \hy) -- ({((14*\res + 8*\gs + 5*\gl)/\tp)*\columnwidth}, \hy) node {FCN+S};
\draw [line width=\lw,  black] ({((14*\res + 8*\gs + 6*\gl)/\tp)*\columnwidth}, \hy) -- ({((15*\res + 8*\gs + 6*\gl)/\tp)*\columnwidth}, \hy) node {FCN};

    \begin{scope}
        \node[anchor=north west, inner sep=0] (image) at (0,0) {\includegraphics[width=1.0\columnwidth]{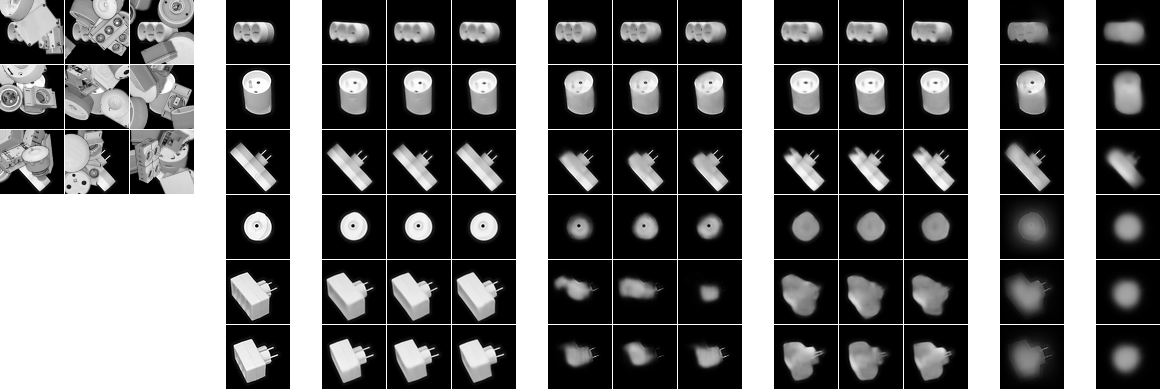}};
    \end{scope}
    \end{tikzpicture}
    \end{minipage}
    \caption{Reconstruction results on \ftless.}
    \label{fig_ftless_reconstruction}
\end{figure}

\cref{fig_fmnist_reconstruction,fig_fceleba_reconstruction,fig_ftless_reconstruction} visualize the reconstruction outputs for all our datasets and architectures. For these results, the target image is always given as input. The first three rows of each figure show the reconstruction, when additionally three noisy or occluded input images are fed into the network.

The images show that our \fvae reconstructs the target images almost perfectly for all three datasets. On \fmnist, only the FCN does not manage to reconstruct the target images but shows blurry versions of them. We also see the same behavior for \fceleba and \ftless which underlines the importance of skip connections for this type of network. On \fceleba, we see that CVAE+S suffers from numeric instabilities causing colorful artifacts in some images. Omitting the skip connections here avoids that issue. On \ftless, all baseline methods create more or less blurry versions of the target image when just the target image is given. When inputting the occluded images in addition to the target image, the reconstruction is much better which shows that these networks have over-fitted to the task of removing occluded objects so that they cannot deal well with non-occluded images. In contrast, \fvae has the ability to reconstruct non-occluded input images very well.

\end{document}